\def\year{2021}\relax
\documentclass[letterpaper]{article} % DO NOT CHANGE THIS
\usepackage{aaai21}  % DO NOT CHANGE THIS
\usepackage{times}  % DO NOT CHANGE THIS
\usepackage{helvet} % DO NOT CHANGE THIS
\usepackage{courier}  % DO NOT CHANGE THIS
\usepackage[hyphens]{url}  % DO NOT CHANGE THIS
\usepackage{graphicx} % DO NOT CHANGE THIS
\def\UrlFont{\rm}  % DO NOT CHANGE THIS
\usepackage{graphicx}  % DO NOT CHANGE THIS
\usepackage{natbib}  % DO NOT CHANGE THIS AND DO NOT ADD ANY OPTIONS TO IT
\usepackage{caption} % DO NOT CHANGE THIS AND DO NOT ADD ANY OPTIONS TO IT
\usepackage{ulem}
\usepackage{hyperref}
\title{Anticipatory Thinking Challenges in Open Worlds: Risk Management}

 \author {
Adam Amos-Binks, \textsuperscript{\rm 1}
Dustin Dannenhauer, \textsuperscript{\rm 2}
Leilani H. Gilpin \textsuperscript{\rm 3}
}
\affiliations {

\textsuperscript{\rm 1} Applied Research Associates, Raleigh, NC 27616 \\
\textsuperscript{\rm 2} Parallax Advanced Research, Beavercreek, OH 45431 \\
\textsuperscript{\rm 3} University of California, Santa Cruz, Santa Cruz, CA 95060 \\
aamosbinks@ara.com, dustin.dannenhauer@parallaxresearch.org, lgilpin@ucsc.edu
}
\usepackage{todonotes}

\begin{document}

\maketitle
%%%%%%%%%%%%%%%%%%%%%%%%%%%%%%%%%%%%%%%%%%%%%%%%%%%%%
% \begin{abstract}
% \adam{Schedule
% \begin{itemize}
%     % \item Web Nov 17: Outline (Adam)
%     % \item Thurs Nov 18: Intro/Conclusion bullets, Fig 1 draft 
%     % \item Fri Nov 19: Intro/Conclusion narrative, Fig 1 final; PC bullets Fig 2 draft
%     % \item Mon Nov 22: PC narrative Fig 2 final; CC bullets Fig 3 draft
%     \item Tues Nov 23: CC narrative Fig 3 draft (by noon ET); Meet and assign final changes
%     \item Wed Nov 24 COB PT: Submit here: (select 'Designing Artificial Intelligence for Open Worlds' as the relevant track) https://easychair.org/conferences/?conf=sss22 
% \end{itemize}
% }
% \end{abstract}
%%%%%%%%%%%%%%%%%%%%%%%%%%%%%%%%%%%%%%%%%%%%%%%%%%%%%
\section{Introduction}\label{sec:intro}
%Using CARS model for intro https://libguides.usc.edu/writingguide/CARS
%%%%%%%%%%%%%%%%%%%%%%%%%%%%%%%%%%%%%%%%%%%%%%%%%%%%%

Anticipatory thinking ~\cite{Geden2019} drives our ability to manage risk -- identification and mitigation -- in everyday life, from bringing an umbrella when it might rain to buying car insurance. As AI systems become part of everyday life, they too have begun to manage risk. Autonomous vehicles log millions of miles \cite{RR-1478-RC}, StarCraft and Go agents have similar capabilities to humans, implicitly managing risks presented by their opponents. To further increase performance in these tasks, out-of-distribution evaluation has emerged as a way to characterize a model's bias, what we view as a type of risk management.

However, learning to identify and mitigate low-frequency, high-impact risks is at odds with the observational bias required to train machine learning models. StarCraft and Go are closed-world domains whose risks are known and mitigations well documented, ideal for learning through repetition. Adversarial filtering datasets provide difficult examples but are laborious to curate and static ~\cite{Hendrycks}, both barriers to real-world risk management. Adversarial robustness focuses on model poisoning under the assumption there is an adversary with malicious intent, without considering naturally occurring adversarial examples. Adversarial generation focuses at the object level (e.g. ~\citealt{Zhao2018}) without an open-world context where adversarial scenes occur. These methods are all important steps towards improving risk management but they do so without considering open-worlds, where new risks and new mitigations require dynamic risk management.

We unify these open-world risk management challenges with two contributions. The first is our perception challenges, designed for agents with imperfect perceptions of their environment whose consequences have a high impact (e.g. autonomous vehicles and public safety). Our second contribution are cognition challenges, designed for agents that must dynamically adjust their risk exposure as they identify new risks and learn new mitigations. Our goal with these challenges is to spur research into solutions that assess and improve the anticipatory thinking required by AI agents to manage risk in open-worlds and ultimately the real-world.

%%%%%%%%%%%%%%%%%%%%%%%%%%%%%%%%%%%%%%%%%%%%%%%%%%%%%
\section{Open World Risk Management}\label{sec:ow_rm}
%%%%%%%%%%%%%%%%%%%%%%%%%%%%%%%%%%%%%%%%%%%%%%%%%%%%%

\begin{figure}[t]
  \centering
  \includegraphics[width=\columnwidth]{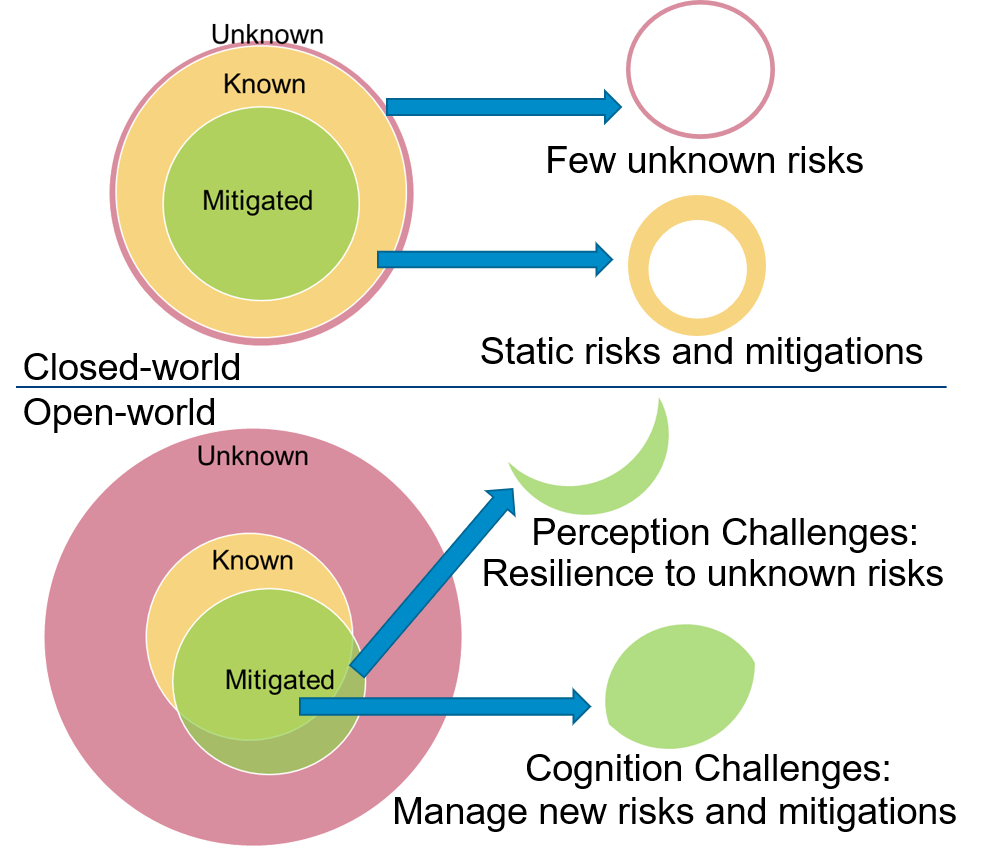}
  \caption{Managing risk in open-worlds is more complex than in closed-worlds. Agents must be resilient to unknown risks and once observed, new risks need to be balanced with existing risks by applying new mitigations.}
  \label{fig:overall_venn}
\end{figure}

\begin{figure*}
\includegraphics[width=\textwidth]{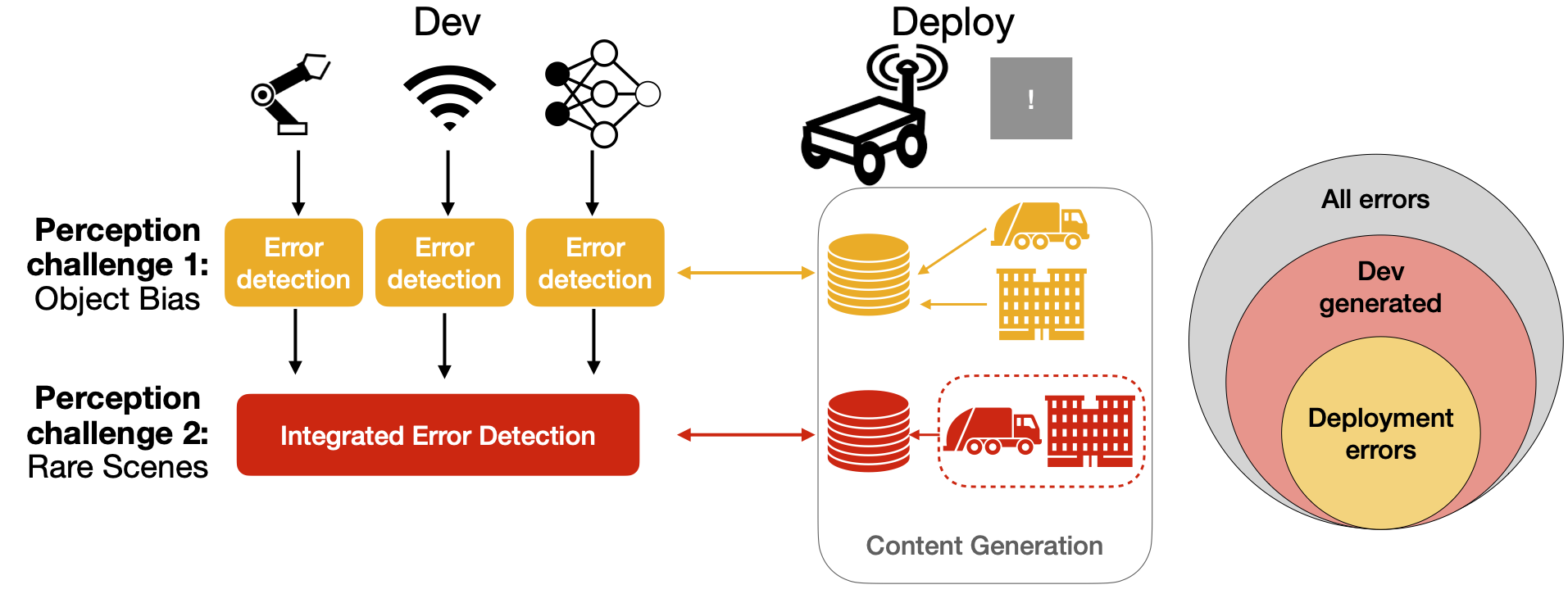}
\caption{Flow diagram of the Perception challenges.  Perception challenge 1: Object bias adds a content generation step at the perception system level.  The inputs to the content generation step are generated from deployment (real, rare objects that were previously incorrectly labelled).  Perception challenge 2: Rare Scenes adds a content generation step at the system level.  The inputs to this content generation step are generated in development; inspired by ``real'' rare scenes; e.g., a truck carrying traffic lights.  }
\label{fig:av-challenge}
\end{figure*}

We view current AI systems like insurance agents that manage risks in a closed-world. Within an insurance domain (e.g. home) there are few unknown risks to insure against and the known risks are static (e.g. flood, fire, etc), see top Figure \ref{fig:overall_venn}. Agents calculate mitigations (premiums) from an ample supply of historical likelihood and impact data. An individual manages their risk by purchasing policy premiums depends on their risk tolerance and budget. In contrast, the open-world perception challenge captures how well a perception system adapts to the disproportional (and growing) number of unknown risks in relation to known risks, see bottom Figure \ref{fig:overall_venn}. For example, perception systems in autonomous vehicles have caused accidents by mis-classifying concrete barriers as the horizon. This epitomizes the often referred `long-tail' of errors, the infrequent but highly impactful situations that open-world AI systems must manage the risk of. Solutions to the perception challenge will provide accurate assessments of a model's ability to identify and manage unknown risks both at the object and scene level. Our cognition challenge assumes high quality perceptions (e.g. identifying new known risks) and focuses on balancing newly identified risks and learning new mitigations to maintain a desired risk profile. Open-world games (those with large, complex action models and catastrophic consequences such as permanent death) are ideal domains as long-range credit assignment is difficult to model and new adversaries and capabilities are constantly being discovered. The following sections detail these challenges in concrete domains and propose example high-level solutions that combine both symbolic reasoning and machine learning to operationalize risk management in open-worlds.

%%%%%%%%%%%%%%%%%%%%%%%%%%%%%%%%%%%%%%%%%%%%%%%%%%%%%
\section{Perception Challenges} \label{sec:perception}
%%%%%%%%%%%%%%%%%%%%%%%%%%%%%%%%%%%%%%%%%%%%%%%%%%%%%
% \adam{My thoughts on what should be here:
% \begin{itemize}
%     \item \textit{Intro autonomous vehicles} - argue self-driving cars are open world because of they must constantly perceive surroundings, must manage risk of imperfect perceptions, new combinations which mean it is open-world. Imperfect perceptions challenges come in at least two forms. 3rd wave approaches provide solutions. 
%     \item \textit{Figure} - extend existing figures we've done to capture both examples below
%     \item \textit{Challenge 1 - "natural adversarial objects"} - Some risk management assessment using out-of-distribution (OOD) datasets (adversarial filtering) but are static and will lead to improvements only on this set. Need generative model informed by cognitive model to truly assess perception system observational bias.
%     \item \textit{Challenge 2 - "natural adversarial scenes"} - Assuming accurate object perception can still be fooled by rare scenes (e.g. traffic light on truck). Again, need generative model informed by cognitive model to generate OOD scenes to truly assess perception system observational bias.
%     \item \textit{Impact} - Addressing these 2 challenges will improve safety, regulation, etc.
% \end{itemize}
% }

Autonomous vehicles are constantly perceiving their surrounding environments with perception systems, e.g., vision, LiDAR, and radar, to resolve the difference between reality and their representation of it.  The vision system, which is opaque to humans, is not prepared for rare but highly impactful perception errors; making autonomous vehicles an open-world domain~\cite{Langley_2020}.

%Autonomous vehicles are complex systems that fail in complex ways.  
One of the challenges of quantifying the risk of autonomous vehicles in the real world is that they are deployed in an open-world but tested in closed-world simulations. This juxtaposition highlights how unprepared autonomous vehicles are for naturally occurring adversarial examples. In this challenge, we suggest an alternative: generating naturally occurring adversarial examples from ``real'' rare cases to supplement the closed-world test simulations, mimicking the open-worlds where they are deployed. We further refine the perception challenge at the object and scene level, summarizing them in Figure~\ref{fig:av-challenge}.

\subsection{Perception Challenge 1: Object Bias}
\label{sec:av-challenge-1}
The current state-of-the-art vision systems are easily fooled~\cite{fooled} by out-of-distribution inputs that exploit a model's observational bias.  Some solutions include training on adversarial examples~\cite{adversarial-training}, but they only lead to improvements on this  data set.  Instead, we propose iterative testing where objects mislabelled in deployment are used as input to a content generation model for testing and evaluation of a perception system's observational biases. The generative model creates sets of naturally adversarial objects that can be tested during development (e.g. faded stop signs, shadow patterns).

%Imperfect perception comes in at least two forms; at the object level and at the scene level.  At the object level, there is an \textit{observational bias}; computer vision object detection algorithms can only detect objects they have been trained on before.  Some risk management assessments use out-of-distribution (OOD) data sets, e.g., adversarial filtering.  But this process is static; it is a one-off fix for a single data set.  Instead, we need an iterative approach to assessing and stress testing object detection systems to truly assess the observational bias, in Challenge 1: Object Bias.

\begin{figure*}
\includegraphics[width=\textwidth]{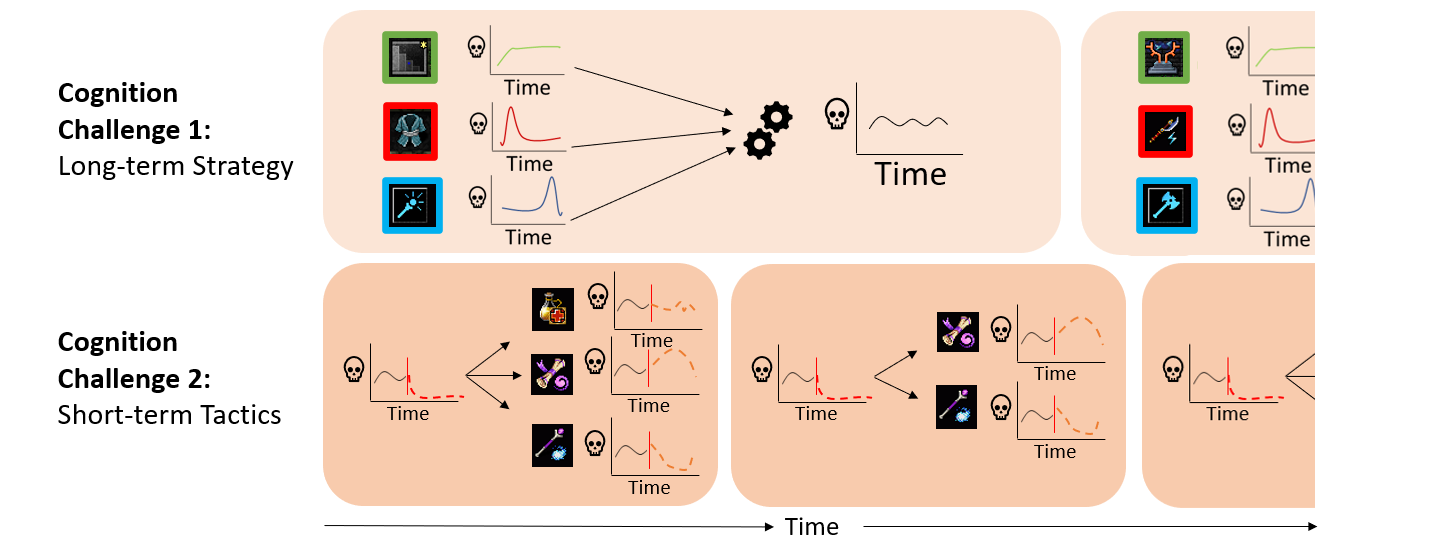}
\caption{We operationalized cognition challenges in the Dungeon Crawl Stone Soup environment that include long-term strategy (top) and short-term tactics (bottom) risk management tasks.}
\label{fig:cc-challenge}
\end{figure*}

\subsection{Perception Challenge 2: Rare Scenes}
\label{sec:av-challenge-2}

A second type of imperfect perception occurs when perceiving rare scenes.  While there is still a potentially problematic observational bias at the object level, we assume that the object detection system is mostly accurate. A vision system can still be ``fooled'' by rare scenes such as traffic lights on a truck\footnote{https://futurism.com/the-byte/tesla-autopilot-bamboozled-truck-traffic-lights}.  In these cases, the perception is correct: the objects are correctly identified, but the scene is so rare and disruptive that it should be immediately recognized so the system can avoid catastrophic failures (e.g. immediately stopping on a freeway). These ``naturally-occurring adversarial scenes'' exemplify the problem with the long tail of errors in autonomous vehicles. An additional layer of reasoning is required to test and monitor for these types of rare scenes.

To address rare scene limitations, we propose a similar iterative process: a generative model that constructs out-of-distribution scenes with commonly occurring objects. The generative model uses objects that are regularly seen in the data (e.g., road signs, objects, etc.) and changes their orientation (e.g., fallen signs on the ground) or environment (e.g., objects in the same scene that usually are not together).

\subsection{Example Solution}
Autonomy requires adapting to operational circumstances that cannot be enumerated pre-hoc. However, assessing this ability is static and contrived, relying on benchmark tasks (e.g. avoid collisions in a benchmark scenario) that systems could be gaming with non-generalizable rules. Methods such as variational auto-encoders (VAEs) and generative adversarial networks (GANs) can generate a whole family of tasks and push the boundaries of contrived task-oriented evaluations to rich attribute-oriented evaluations. We envision integrating physical world knowledge into content generation methods to ensure real-world properties are preserved and natural adversarial instances are generated. For example, static traffic indicators (stop signs, traffic lights, etc) that are now dynamic (e.g. loaded onto a flatbed truck, advertisement painted on a large truck). Performance against a family of out-of-distribution tasks would indicate a system's competency in a risk management and provide a level of insight into future performance not currently available.

% Took this from the previous paper.  
\subsection{Impact}
\label{sec:perception-impact}
Autonomous vehicles in new, open environments are susceptible to causing harm (even death). Despite high accuracy in testing, object recognition systems cannot account for rare but consequential out-of-distribution inputs. Testing and improving a perceptions system's ability to manage the risk presented by these new circumstances would make autonomous systems safer and cover a larger number of error and failure cases. This type of anticipatory, introspective, solution impacts public safety, manufacturers, and regulators.
% Need one more sentence...

%%%%%%%%%%%%%%%%%%%%%%%%%%%%%%%%%%%%%%%%%%%%%%%%%%%%%
\section{Cognition Challenges}\label{sec:eval}
%%%%%%%%%%%%%%%%%%%%%%%%%%%%%%%%%%%%%%%%%%%%%%%%%%%%%

% \adam{My thoughts on what should be here:
% \begin{itemize}
%     \item \textit{Intro DCSS} - argue DCSS environment is open world because of level/adversary/capability scale, need to adapt and managing risks in newly discovered world properties. Managing risks challenges come in at least two forms. 3rd wave approaches provide solutions. 
%     \item \textit{Figure} - extend existing figures we've done to capture both examples below
%     \item \textit{Challenge 1 - "pre-deployment stage"} - Manage risk of actions with long-range dependencies (credit assignment). Need cognitive model to truly assess perception system observational bias.
%     \item \textit{Challenge 2 - "deployment stage"} - Discover new action models with new risks, decide when to learn new things, learning strategy
%     \item \textit{Impact} - Addressing these 2 challenges will improve autonomy
% \end{itemize}
% }

Our cognition challenges for anticipatory thinking in open worlds involve AI agents acting in unexplored, hostile, and dynamic environments. The first challenge: \textit{long-term strategy} concerns issues of risk as agents move between situations and micro environments where resource availability changes. The second challenge: \textit{short-term tactics} concerns more immediate risks that require agents to make use of the right resources at the right time and choice of enemy engagements. We use the rogue-like video game Dungeon Crawl Stone Soup (DCSS) to operationalize this challenge problem. In DCSS a player moves through a procedurally generated, partially observable, and stochastic environment to retrieve the `Orb of Zot' while managing the risks (permanent death) of encountering thousands of monsters. DCSS is one of two rogue-like games with increased interest in recent years and remains an unsolved domain for AI \cite{dannenhauer2021dcss, kuettler2020nethack}.

We consider DCSS an open-world for both human and AI agents owing to over 600 unique monsters and 100 item types that a player comes across as they delve deeper into the 100 levels of the dungeon. As the player enters new areas of the game, the chances of encountering each type of monster and item change, such that certain items and monsters cannot be found until reaching a certain depth. Therefore as a player who begins after only playing the short tutorial, there are many surprises and unknowns encountered. For this reason, DCSS can be considered an open-world for any learning agent who does not start with a complete model of the environment (such a model likely only exists in the minds of the lead developers). For use in AI research, \textit{dcss-ai-wrapper} \cite{dannenhauer2021dcss} provides an API for both symbolic and vector AI agents to interact with DCSS.

\subsection{Cognition Challenge 1: Long-term Strategy}
To win a game of DCSS, AI agents must mitigate risk over longer time horizons as they visit different regions of a world. Humans perform this type of reasoning frequently, such as going from home to work, going from the grocery store to a camping trip, or going from the airport lobby through security to their gate. In all cases, agents are able to take actions in the source region that may no longer be available in the destination region. Such actions could include whether to pack an umbrella and lunch box, or taking a water filter. These same challenges are present in the DCSS domain and require decision making to decide where in the dungeon to go next and choices in enhancing character capabilities. The effects of these decisions are often noticed only after significant time delays as shown in the top portion of Figure \ref{fig:cc-challenge}. We now highlight two aspects of the long-term strategy challenge in DCSS:\\

\noindent\textbf{Capability Enhancement via Skill Point Allocation:} The player earns experience when killing monsters which is permanently allocated to one or more of 33 skills. These skills impact the success of a variety of player-actions and defensive capabilities. Failure in the mid to late game often results from poor risk management of skill points.\\

\noindent\textbf{Region Transitions via Level Choices:} There are 24 themed branches (regions) of the dungeon, each with their own special characteristics. Some branches are known for an overwhelming majority of poison-based monsters (spider lair and snake pit), corrosive monsters (slime pit), monsters that will cause unhelpful mutations to your character (the abyss), or monsters with special abilities, such as locking stairs preventing escape (the vaults). Further increasing complexity, monsters move in levels independent of your movement, it is likely that as you explore a level, monsters will move in between your planned escape route from the location at which you entered the level. Entering a branch without dynamically adjusting risks and mitigations (poison resist gear, corrosion resist gear, potions of cure mutation, etc.) will likely lead to catastrophic failure.

\subsection{Cognition Challenge 2: Short-term Tactics}
The limited visibility and highly dynamic, stochastic nature of item and monster interactions means that there are many possible futures over short horizons consisting of a small ($< 5$) possible actions. Additionally, agents may never encounter uncommon situations more than once because of the procedurally generated nature of the game. Therefore short-term anticipatory thinking should seek to mitigate the absolute worst outcomes even if unable to accurately predict a single outcome. These types of decisions are shown in the bottom portion of Figure \ref{fig:cc-challenge}. We now highlight two aspects of the short-term tactics challenge in DCSS:\\

\noindent\textbf{Engagements via Direct Combat:} It is impossible to progress into the mid to late game by continuously engaging in combat with little thought to tactical reasoning. Retreating, hiding from monsters, using special abilities, and using other terrain features such as hallways, are needed to gain advantages (i.e. in a hallway you can fight monsters in smaller batches). \\

\noindent\textbf{Resource Management via Item Inventory:} The player will encounter tens of thousands of item instances from ~100 item types and the player has a limited inventory space. Sometimes items may need to be left and returned to later, other items dropped to make sure the most important life saving items are kept for dire situations. Additionally, the choice to use or save a life-saving consumable item is often the difference between success and failure. These types of decisions must be made many times during gameplay.

\subsection{Example Solutions}

Simply learning a complete action model of an open-world is intractable and so is encoding all knowledge an agent needs for managing risks. A hybrid or 3rd wave approach would reason over the semantics of what new capabilities or assets would produce a desired risk profile  (e.g. goal reasoning, meta-cognition), taking into account the small samples of newly observed risks (e.g. meta-learning, few-shot learning). This approach is just an illustrative example of the complexities required to go beyond single task performance and manage risk in open-worlds.

\subsection{Impact}
Developing AI agents that can solve these two challenges in DCSS will lead to approaches that can lead to better AT in a wide variety of autonomous systems. An autonomy approach that can solve the right choice of entering which branch of the dungeon could be the same approach that helps an autonomous drone decide whether it can follow a target into a broken down urban building, a forest, or a cave.  An AI that is tasked with it's mission and can solve the pre-deployment problem could alert it's operating before it leaves the forward operating base that it's missing a needed component to mitigate a possible risk from adversaries that the operator may have forgotten. An AI agent that can manage risks during operation will be able to make the call that it cannot outrun a new enemy and must seek to hide instead.

%%%%%%%%%%%%%%%%%%%%%%%%%%%%%%%%%%%%%%%%%%%%%%%%%%%%%
\section{Conclusion}\label{sec:conclusion}
%%%%%%%%%%%%%%%%%%%%%%%%%%%%%%%%%%%%%%%%%%%%%%%%%%%%%
Real-world deployments of AI systems require a level of robustness and safety best tested in open-worlds. We use a risk management approach to contribute two challenges and illustrate them in different domains. Our perception challenges characterize a systems ability to recognizing both rare objects and configurations of them. Solutions with this capability are essential as the real-world is constantly producing new permutations of objects and contexts they appear in. Our cognition challenges characterize the relationship between new risks and new capabilities to mitigate them. Dynamically balancing a risk profile with new information enables AI agents to adapt to situations like humans do and avoid needless catastrophic failures. Our example solutions focus on integrating symbolic and learning approaches together to manage risks in perception and cognition. 
%Solutions to these challenges will push assessments and capabilities to high-assurance levels of robustness and safety.

%%%%%%%%%%%%%%%%%%%%%%
%\section{Acknowledgements}
%This study is supported by XXX. The views expressed in this paper are those of the authors and do not necessarily represent the official position or policy of the U.S. Government, or the Department of Defense. 

%\bibliographystyle{aaai21}
\cleardoublepage
\bibliography{bib}

\end{document}